\documentclass[runningheads]{llncs}
\usepackage{graphicx}
\usepackage{grffile}
\usepackage{amsmath,amssymb}
\usepackage{xcolor}
\usepackage{hyperref}
\usepackage{booktabs}
\usepackage{multirow}
\title{Unified Framework for Tabular Data Generation:\\From GANs to Diffusion Models and Large Language Models}
\author{Insaf Ashrapov\inst{1}\orcidID{0000-0003-4938-0430} }

\institute{Moscow Institute of Physics
and Technology, Moscow, Russia
\email{intoff@mipt.ru} \url{https://eng.mipt.ru/} }

\begin{document}

\maketitle

\begin{abstract}
Generative models for tabular data have evolved rapidly beyond Generative Adversarial Networks (GANs). While GANs pioneered synthetic tabular data generation, recent advances in diffusion models and large language models (LLMs) have opened new paradigms with complementary strengths in sample quality, privacy, and controllability. In this paper, we survey the landscape of tabular data generation across three major paradigms---GANs, diffusion models, and LLMs---and introduce a unified, modular framework that supports all three. The framework encompasses data preprocessing, a model-agnostic interface layer, standardized training and inference pipelines, and a comprehensive evaluation module. We validate the framework through experiments on seven benchmark datasets, demonstrating that GAN-based augmentation can improve downstream performance under distribution shift. The framework and its reference implementation are publicly available at \url{https://github.com/Diyago/Tabular-data-generation}, facilitating reproducibility and extensibility for future research.
\end{abstract}


%%=============================================================================
\section{Introduction}\label{sec:introduction}
%%=============================================================================

Tabular data remains the most prevalent data modality in industry and scientific applications, underpinning domains from healthcare and finance to e-commerce and social science~\cite{borisov2022deep}. Yet acquiring sufficient high-quality labeled tabular data is often expensive, privacy-constrained, or limited by distributional shift between training and deployment environments. Data stability is critical in applied domains such as geophysical segmentation~\cite{karchevskiy2018salt} and satellite-based construction monitoring~\cite{ashrapov2022satellite}, where distribution shift directly impacts model reliability. Synthetic data generation addresses these challenges by learning the joint distribution of a table and producing new rows that preserve statistical fidelity while mitigating privacy risks.

Generative Adversarial Networks (GANs)~\cite{goodfellow2014generative} were among the first deep generative models adapted for tabular data~\cite{xu2018synthesizing,xu2019modeling}. Their adversarial training paradigm has proven effective but suffers from well-known issues such as mode collapse, training instability, and difficulty handling mixed data types. More recently, \emph{denoising diffusion probabilistic models}~\cite{ho2020denoising} have demonstrated strong performance in image synthesis and are increasingly being adapted to structured data~\cite{kotelnikov2023tabddpm,zhang2023mixed}. Concurrently, \emph{large language models} (LLMs) have shown surprising efficacy at generating tabular rows by treating them as serialized text~\cite{borisov2023language,solatorio2023realtabformer}. These three paradigms---GANs, diffusion models, and LLMs---each offer distinct trade-offs in sample quality, training efficiency, controllability, and privacy guarantees.

Despite this progress, the field lacks a unified software abstraction that enables practitioners to compare, combine, and deploy these heterogeneous approaches within a single experimental pipeline. Existing work typically presents isolated method-specific implementations, making fair comparison and reproducibility difficult.

\paragraph{Contributions.} The main contributions of this work are as follows:
\begin{enumerate}
    \item We provide a comprehensive survey of tabular data generation spanning GANs, diffusion models, LLMs, and hybrid approaches, with emphasis on advances from 2023--2025.
    \item We propose a \emph{unified, modular framework} for tabular data generation that abstracts over multiple generative paradigms through a common model interface, standardized preprocessing, and a unified evaluation module. The framework is designed to be extensible to future generative paradigms.
    \item We release an open-source reference implementation of the framework at \url{https://github.com/Diyago/Tabular-data-generation}, serving as the experimental backbone and reproducibility layer for all results reported in this paper.
    \item We conduct experiments on seven benchmark datasets demonstrating the utility of GAN-based augmentation under distribution shift, with all experiments built on top of the proposed framework.
\end{enumerate}

The remainder of this paper is organized as follows. Section~\ref{sec:background} reviews the foundations of generative models. Section~\ref{sec:tabular_generation} surveys tabular data generation across paradigms. Section~\ref{sec:framework} presents the proposed unified framework. Section~\ref{sec:experiments} describes our experimental evaluation. Section~\ref{sec:discussion} discusses data quality, privacy, controllability, and scalability. Section~\ref{sec:conclusion} concludes.

%%=============================================================================
\section{Background: Generative Models}\label{sec:background}
%%=============================================================================

\subsection{Generative Adversarial Networks}\label{sec:gan_background}

\begin{figure}[!htbp]
\centering
\includegraphics[width=0.9\columnwidth]{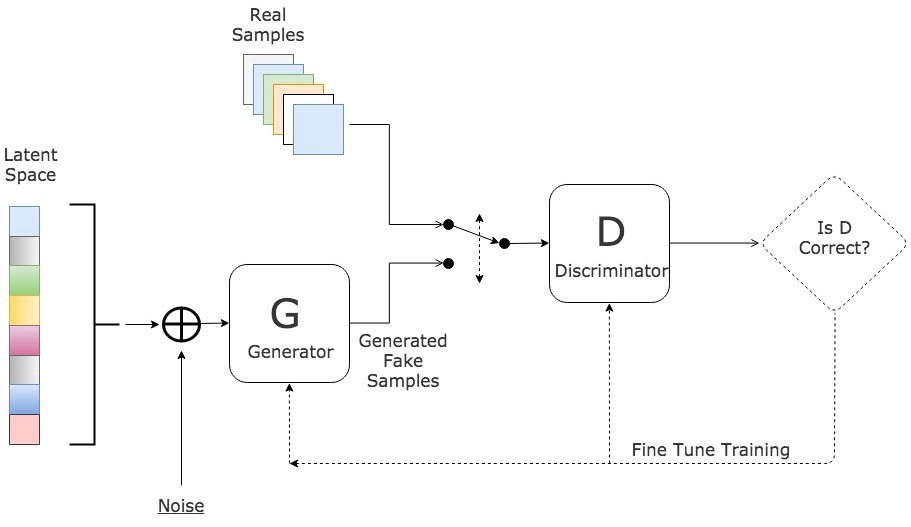}
\caption{GAN training pipeline~\cite{wutgan}. The generator produces synthetic samples from random noise, and the discriminator learns to distinguish real from fake data.}
\label{generator}
\end{figure}

\begin{figure}[!htbp]
\centering
\includegraphics[width=0.9\columnwidth]{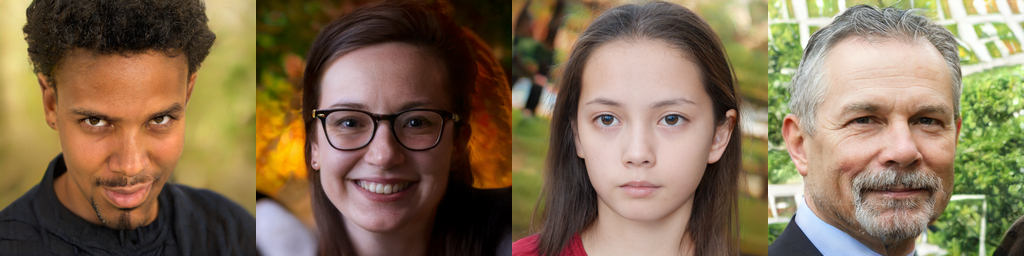}
\caption{Photo-realistic human faces generated by StyleGAN~2~\cite{karras2019analyzing}.}
\label{faces}
\end{figure}

\begin{figure}[!htbp]
\centering
\includegraphics[width=0.9\columnwidth]{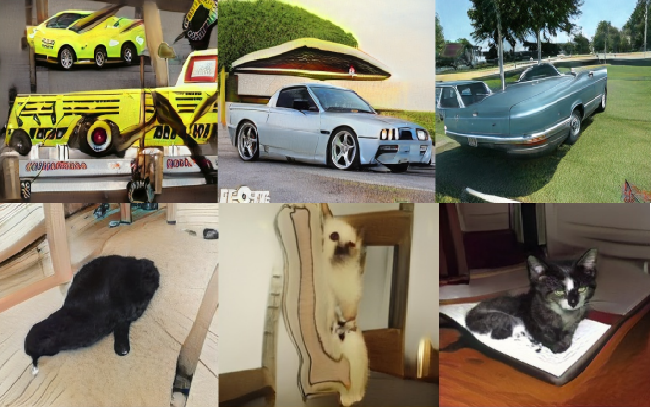}
\caption{Failure cases of StyleGAN~2 on cars and cats~\cite{karras2019analyzing}, illustrating remaining challenges in complex object generation.}
\label{cats}
\end{figure}

A GAN~\cite{goodfellow2014generative} consists of two neural networks trained simultaneously: a \textbf{generator} $G$ that maps latent noise $z \sim p_z$ to synthetic samples, and a \textbf{discriminator} $D$ that distinguishes real from generated samples. The training objective is a minimax game:
\begin{equation}
\min_G \max_D \; \mathbb{E}_{x \sim p_{\text{data}}}[\log D(x)] + \mathbb{E}_{z \sim p_z}[\log(1 - D(G(z)))].
\end{equation}
The general architecture and training pipeline are illustrated in Figure~\ref{generator}. Modern GAN variants such as StyleGAN~2~\cite{karras2019analyzing} can produce photo-realistic images (Figure~\ref{faces}), though challenges remain with complex scenes (Figure~\ref{cats}).

Key challenges with GANs include:
\begin{itemize}
    \item \textbf{Training cost.} State-of-the-art architectures require substantial compute (e.g., one week on 8$\times$ NVIDIA Tesla V100 for StyleGAN~2).
    \item \textbf{Mode collapse.} The generator may fail to cover all modes of the data distribution.
    \item \textbf{Training instability.} Balancing generator and discriminator learning rates requires careful tuning.
\end{itemize}

\subsection{Denoising Diffusion Probabilistic Models}\label{sec:diffusion_background}

Diffusion models~\cite{ho2020denoising,song2021scorebased} define a forward process that gradually adds Gaussian noise to data over $T$ steps, and learn a reverse denoising process to generate samples. The forward process is defined as:
\begin{equation}
q(x_t | x_{t-1}) = \mathcal{N}(x_t; \sqrt{1-\beta_t}\, x_{t-1},\; \beta_t \mathbf{I}),
\end{equation}
where $\{\beta_t\}_{t=1}^T$ is a noise schedule. The reverse process is parameterized by a neural network $\epsilon_\theta$ trained to predict the added noise:
\begin{equation}
\mathcal{L} = \mathbb{E}_{t, x_0, \epsilon}\left[\|\epsilon - \epsilon_\theta(x_t, t)\|^2\right].
\end{equation}
Diffusion models avoid the adversarial training dynamics of GANs and typically exhibit better mode coverage, at the cost of slower sampling due to iterative denoising.

\subsection{Large Language Models}\label{sec:llm_background}

Large language models (LLMs) such as GPT-family models~\cite{brown2020language} are autoregressive transformers trained on massive text corpora. By serializing tabular rows into structured text (e.g., ``\texttt{age: 35, income: 50000, city: Moscow}''), LLMs can generate new rows through next-token prediction. Key mechanisms include:
\begin{itemize}
    \item \textbf{Schema-to-text serialization}: Converting column names and values into natural language or structured string representations.
    \item \textbf{Prompt-based generation}: Providing few-shot examples of table rows as context for in-context learning.
    \item \textbf{Instruction tuning}: Fine-tuning LLMs on tabular generation tasks with explicit instructions about column semantics, constraints, and distributions.
\end{itemize}
The advantage of LLM-based approaches is their ability to leverage pre-trained world knowledge about feature semantics and inter-column relationships, without task-specific architectural modifications.

%%=============================================================================
\section{Tabular Data Generation}\label{sec:tabular_generation}
%%=============================================================================

While image generation has been the primary showcase for generative models, tabular data presents distinct challenges: heterogeneous column types (numerical, categorical, temporal), complex inter-column dependencies, multi-modal distributions, sparse one-hot encodings, and class imbalance~\cite{xu2018synthesizing}. This section surveys the three major generative paradigms as applied to tabular data.

\subsection{GAN-Based Approaches}\label{sec:gan_tabular}

\subsubsection{TGAN.}
TGAN~\cite{xu2018synthesizing} was among the first architectures specifically designed for tabular data generation. It addresses the heterogeneity problem through specialized preprocessing:

\paragraph{Numerical preprocessing.} Neural networks generate values most effectively within $(-1, 1)$ via $\tanh$, but struggle with multi-modal distributions. TGAN fits a Gaussian Mixture Model (GMM)~\cite{viroli2017deep} with $m{=}5$ components for each continuous column $C$, producing a normalized value $V$ and a cluster probability vector $U$.

\paragraph{Categorical preprocessing.} Categorical variables are converted to one-hot encodings with added noise, and probability distributions are generated via softmax.

\paragraph{Generator.} Numerical variables are generated in two steps: first the value scalar $V$, then the cluster vector $U$ with $\tanh$ activation. Categorical features are generated as probability distributions over labels via softmax. An LSTM~\cite{10.1162/neco.1997.9.8.1735} with attention generates features sequentially.

\begin{figure}[!htbp]
\centering
\includegraphics[width=0.9\columnwidth]{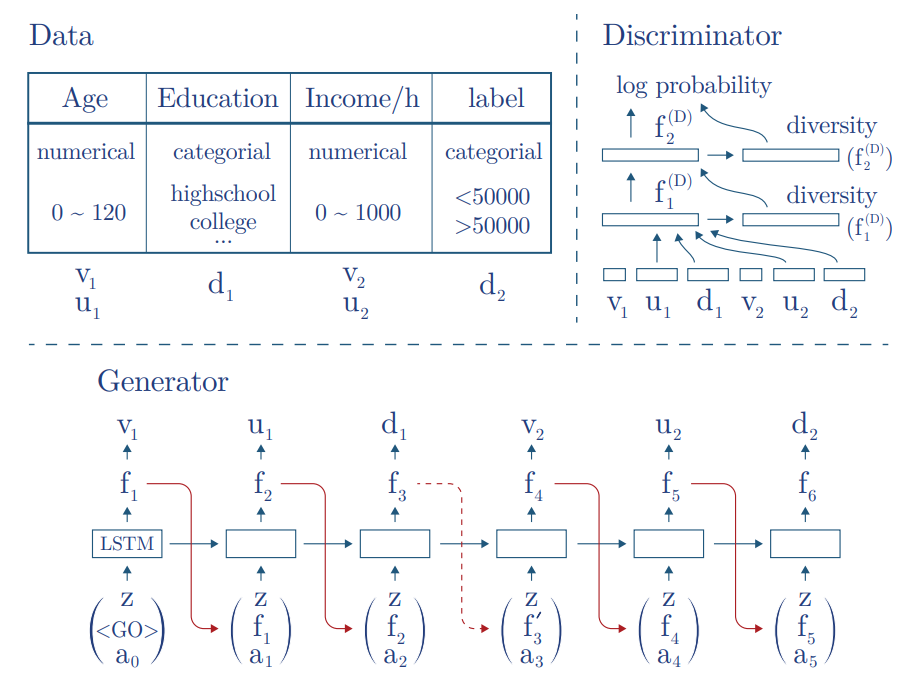}
\caption{TGAN architecture. The generator produces features sequentially using an LSTM. The discriminator concatenates all features and uses an MLP to classify real vs.\ synthetic data~\cite{xu2018synthesizing}.}
\label{disc}
\end{figure}

\paragraph{Discriminator.} A multi-layer perceptron (MLP) with LeakyReLU~\cite{xu2015empirical} and batch normalization~\cite{ioffe2015batch} processes the concatenated feature vectors. The loss combines KL divergence with ordinal log loss. The architecture is shown in Figure~\ref{disc}.

TGAN was evaluated on KDD99 and Covertype datasets, achieving an average performance gap of 5.7\% between models trained on real vs.\ synthetic data~\cite{xu2018synthesizing}.

\subsubsection{CTGAN.}
\begin{figure*}[!htbp]
\centering
\includegraphics[width=0.8\textwidth]{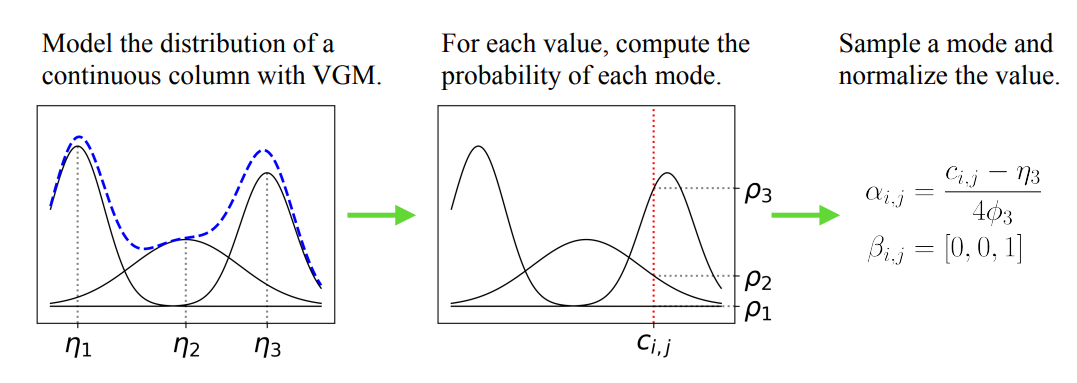}
\caption{Mode-specific normalization in CTGAN~\cite{xu2019modeling}. Continuous values are normalized within each Gaussian mode, enabling effective handling of multi-modal distributions.}
\label{exm}
\end{figure*}

CTGAN~\cite{xu2019modeling} introduced three key improvements over TGAN:

\paragraph{Mode-specific normalization.} Instead of a fixed GMM, CTGAN employs a variational Gaussian mixture model (VGM) that automatically estimates the number of modes $m$. Each continuous value is normalized within its assigned mode, represented as a scalar $\alpha$ and a one-hot mode indicator $\beta$. An example is shown in Figure~\ref{exm}.

\paragraph{Conditional generator.} A conditional vector encodes the selected category across all discrete columns. For instance, given $D_1 = \{1,2,3\}$ and $D_2 = \{1,2\}$, the condition $D_2=1$ yields mask vectors $\mathbf{m}_1 = (0,0,0)$ and $\mathbf{m}_2 = (1,0)$, so $\text{cond} = (0,0,0,1,0)$. The generator loss is augmented with a cross-entropy penalty encouraging the generated discrete values to match the conditioning mask.

\begin{figure}[!htbp]
\centering
\includegraphics[width=0.9\columnwidth]{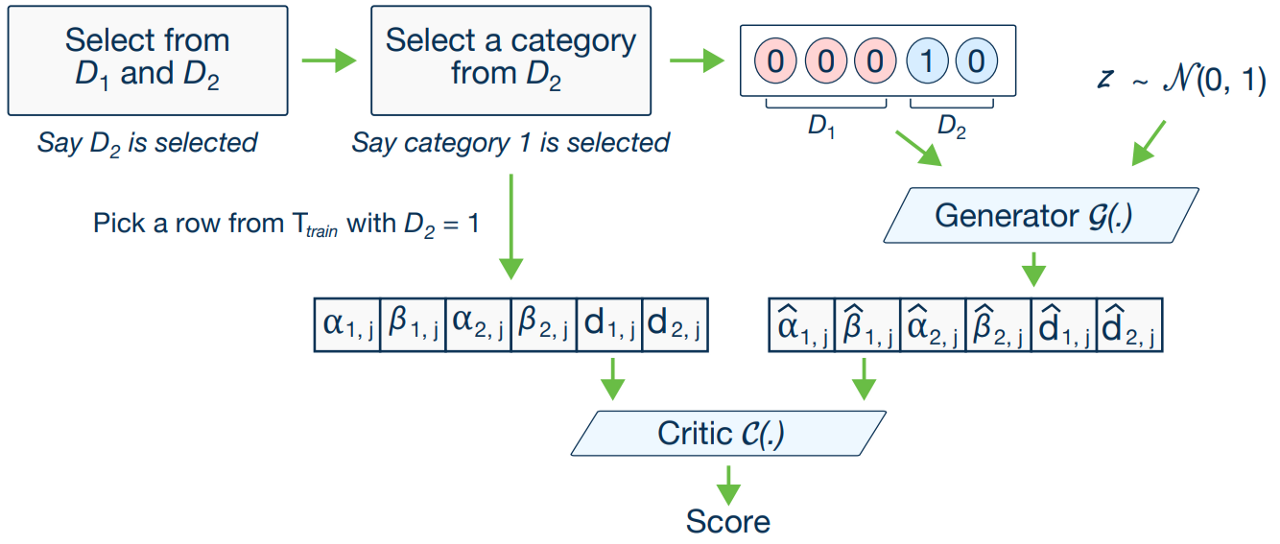}
\caption{CTGAN model. The conditional generator produces rows conditioned on discrete columns, with training-by-sampling ensuring balanced exploration of categorical values~\cite{xu2019modeling}.}
\label{ctgan}
\end{figure}

\paragraph{Training-by-sampling.} Categories are sampled according to their log-frequency during training, ensuring that the model explores all discrete values evenly. The architecture (Figure~\ref{ctgan}) replaces the LSTM with an MLP and uses WGAN loss with gradient penalty.

\begin{figure}[!htbp]
\centering
\includegraphics[width=0.9\columnwidth]{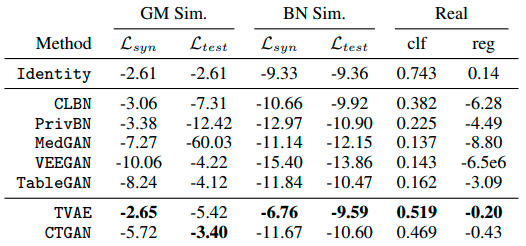}
\caption{Benchmark results from~\cite{xu2019modeling}. CTGAN and TVAE outperform prior methods across simulated and real datasets.}
\label{reults2}
\end{figure}

CTGAN and the companion TVAE model outperform prior methods across Gaussian mixture, Bayesian network, and real-data benchmarks (Figure~\ref{reults2}).

\subsubsection{Recent GAN advances.}
Since CTGAN, several GAN-based improvements have been proposed. Notable directions include regularization techniques to mitigate mode collapse in tabular settings~\cite{zhao2024ctab}, auxiliary classifier mechanisms that improve categorical fidelity, and architectural innovations using transformer-based discriminators. These methods generally retain the CTGAN preprocessing pipeline while improving training stability and sample diversity.

\subsection{Diffusion-Based Approaches}\label{sec:diffusion_tabular}

The adaptation of diffusion models to tabular data has emerged as a promising direction since 2023. The core challenge lies in handling the mixed discrete-continuous nature of tabular features within the continuous diffusion framework.

\paragraph{Multinomial diffusion for categorical features.} Unlike images, tabular data contains categorical columns that cannot be naturally represented in continuous space. Multinomial diffusion processes~\cite{hoogeboom2021argmax} address this by defining forward corruption as transitions between categorical states rather than additive Gaussian noise. During the forward process, categorical values are gradually corrupted toward a uniform distribution over categories, and the reverse process learns to reconstruct the original categories.

\paragraph{Hybrid diffusion architectures.} Recent work proposes handling continuous and categorical columns through separate diffusion processes that are jointly trained~\cite{zhang2023mixed}. Continuous columns follow standard Gaussian diffusion, while categorical columns use multinomial diffusion. A shared denoising network learns cross-column dependencies, enabling coherent row-level generation.

\paragraph{Score-based approaches.} Score-based generative models~\cite{song2021scorebased} estimate the gradient of the log-density (score function) and generate samples via Langevin dynamics. For tabular data, score matching can be applied to continuous features while categorical features are handled through conditional generation or embedding-based approaches.

Diffusion-based methods generally achieve superior mode coverage compared to GANs---the iterative denoising process is less prone to mode collapse---and recent benchmarks~\cite{kotelnikov2023tabddpm} demonstrate competitive or superior downstream utility. However, they incur higher computational cost at inference time due to the multi-step sampling process.

\subsection{LLM-Based Approaches}\label{sec:llm_tabular}

Large language models offer a fundamentally different paradigm for tabular data generation by treating table rows as sequences of tokens.

\paragraph{Row serialization.} Each table row is converted to a text string preserving column names and values, e.g., \texttt{"age is 35, income is 50000, occupation is engineer"}. The serialization format significantly impacts generation quality; structured formats with explicit delimiters and column headers tend to outperform free-form text~\cite{borisov2023language}.

\paragraph{Few-shot and in-context generation.} Pre-trained LLMs can generate new rows given a small number of example rows as prompt context. This enables zero-training-cost generation for small tables, though quality degrades for complex distributions or high-cardinality categorical features.

\paragraph{Fine-tuned tabular LLMs.} More effective approaches fine-tune pre-trained language models on the target table using instruction tuning or causal language modeling objectives~\cite{solatorio2023realtabformer}. The model learns column-specific distributions and inter-column dependencies through the sequential prediction of token sequences representing each row.

\paragraph{Advantages and limitations.} LLM-based generation leverages pre-trained semantic knowledge, enabling meaningful generation even with very limited training data. LLMs naturally accommodate mixed types, as all values are serialized into a common token space. However, they struggle with precise numerical distributions, exhibit high inference costs for large tables, and may hallucinate statistically implausible values. Privacy is also a concern, as pre-trained weights may encode memorized patterns from training corpora.

\subsection{Hybrid Approaches}\label{sec:hybrid}

Recognizing that no single paradigm dominates across all criteria, recent work explores hybrid architectures:
\begin{itemize}
    \item \textbf{GAN--diffusion hybrids}: Using diffusion-based generators with adversarial discriminator losses to accelerate sampling while maintaining sample quality~\cite{xiao2022tackling}.
    \item \textbf{LLM-guided GANs}: Leveraging LLM-generated feature descriptions or semantic embeddings to condition GAN generators, improving categorical coherence.
    \item \textbf{Retrieval-augmented generation}: Combining retrieval mechanisms with generative models, where similar real rows are retrieved to condition generation, improving fidelity for rare patterns.
\end{itemize}

\subsection{Paradigm Comparison}\label{sec:paradigm_comparison}

Table~\ref{tab:paradigm} summarizes the trade-offs across the three major paradigms.

\begin{table}[!htbp]
\centering
\caption{Comparison of generative paradigms for tabular data.}
\label{tab:paradigm}
\resizebox{.95\columnwidth}{!}{
\begin{tabular}{l|c|c|c}
\toprule
\textbf{Criterion} & \textbf{GAN} & \textbf{Diffusion} & \textbf{LLM} \\
\midrule
Training stability & Low & High & High \\
Mode coverage & Moderate & High & Moderate \\
Sample quality & High & High & Moderate \\
Inference speed & Fast & Slow & Moderate \\
Mixed-type handling & Moderate & Moderate & High \\
Privacy guarantees & Low & Moderate & Low \\
Few-shot capability & None & None & High \\
Controllability & Low & Moderate & High \\
Computational cost & Low & High & Very High \\
\bottomrule
\end{tabular}
}
\end{table}

GANs offer fast inference and reasonable quality but suffer from training instability and mode collapse. Diffusion models provide the best mode coverage and training stability at the cost of slow iterative sampling. LLMs excel at controllability and few-shot scenarios but struggle with numerical precision and incur high computational costs. These complementary strengths motivate the unified framework presented in the next section.

%%=============================================================================
\section{Proposed Framework for Tabular Data Generation}\label{sec:framework}
%%=============================================================================

We propose a unified, modular framework for tabular data generation that abstracts over the three generative paradigms discussed above. The framework is designed around four core modules---\emph{data preprocessing}, \emph{model interface layer}, \emph{training and inference pipeline}, and \emph{evaluation module}---connected through standardized interfaces. The reference implementation is publicly available at \url{https://github.com/Diyago/Tabular-data-generation}.

\subsection{Design Principles}\label{sec:design}

The framework is guided by three principles:
\begin{enumerate}
    \item \textbf{Modularity.} Each component (preprocessing, model, training, evaluation) is independently replaceable. A new generative model can be integrated by implementing a single model interface, without modifying preprocessing or evaluation logic.
    \item \textbf{Extensibility.} The framework accommodates future paradigms (e.g., flow matching, energy-based models) through its model-agnostic interface layer. Adding support for a new paradigm requires only implementing the \texttt{fit()} and \texttt{sample()} methods.
    \item \textbf{Reproducibility.} Fixed random seeds, versioned data splits, and standardized evaluation metrics ensure that experiments are fully reproducible. All experimental configurations are stored as declarative specifications.
\end{enumerate}

\subsection{Data Preprocessing Module}\label{sec:preprocessing}

The preprocessing module transforms raw tabular data into representations suitable for each generative paradigm:
\begin{itemize}
    \item \textbf{Numerical features}: Mode-specific normalization via variational Gaussian mixture models (following CTGAN~\cite{xu2019modeling}), standard scaling, or quantile transformation.
    \item \textbf{Categorical features}: One-hot encoding, ordinal encoding, or token-level encoding (for LLM-based models).
    \item \textbf{Row serialization}: For LLM-based generators, the module provides configurable row-to-text serialization with support for multiple formats (key-value pairs, CSV-style, natural language).
    \item \textbf{Missing value handling}: Configurable imputation strategies or explicit missingness indicators.
\end{itemize}
The preprocessing module exposes a uniform API: \texttt{fit\_transform(data)} for training and \texttt{inverse\_transform(synthetic)} for converting generated representations back to the original data schema.

\subsection{Model Interface Layer}\label{sec:model_interface}

The model interface layer defines a common abstraction for all generative models through two core methods:
\begin{itemize}
    \item \texttt{fit(data, config)}: Train the generative model on preprocessed data with the given configuration.
    \item \texttt{sample(n, conditions)}: Generate $n$ synthetic rows, optionally conditioned on specified column values or constraints.
\end{itemize}

This interface currently supports three paradigm-specific backends:
\begin{itemize}
    \item \textbf{GAN backend}: Implements CTGAN-style architectures with configurable generator and discriminator networks, training-by-sampling, and conditional generation.
    \item \textbf{Diffusion backend}: Supports Gaussian diffusion for continuous features and multinomial diffusion for categorical features, with configurable noise schedules and denoising network architectures.
    \item \textbf{LLM backend}: Provides fine-tuning and prompt-based generation interfaces, with configurable serialization formats and decoding strategies (temperature, top-$k$, nucleus sampling).
\end{itemize}

\subsection{Training and Inference Pipeline}\label{sec:pipeline}

The pipeline module orchestrates end-to-end workflows:
\begin{enumerate}
    \item \textbf{Data ingestion}: Load and validate input tables against a user-specified schema.
    \item \textbf{Preprocessing}: Apply the appropriate transformation pipeline based on the selected model backend.
    \item \textbf{Model training}: Train the generative model with configurable hyperparameters, early stopping, and checkpointing.
    \item \textbf{Synthetic data generation}: Produce synthetic tables of specified size, with optional conditioning and post-processing constraints.
    \item \textbf{Post-processing}: Apply inverse transformations and validate generated data against the original schema (type checking, range enforcement, referential integrity for multi-table settings).
\end{enumerate}

For the adversarial augmentation workflow used in our experiments (Section~\ref{sec:experiments}), the pipeline additionally supports: (a)~training a discriminative model to distinguish real from synthetic data, (b)~scoring and filtering synthetic rows by their similarity to a target distribution, and (c)~constructing augmented training sets by combining top-scoring real and synthetic rows.

\subsection{Evaluation Module}\label{sec:evaluation}

The evaluation module provides standardized metrics for assessing synthetic data quality:
\begin{itemize}
    \item \textbf{Statistical fidelity}: Column-wise distributional similarity (Kolmogorov--Smirnov test for continuous, $\chi^2$ test for categorical), pairwise correlation preservation, and mutual information comparison.
    \item \textbf{Machine learning utility (TSTR)}: Train-on-Synthetic, Test-on-Real~\cite{xu2019modeling}---a downstream classifier or regressor is trained on synthetic data and evaluated on held-out real data. Metrics include accuracy, F1, AUC-ROC, and RMSE as appropriate.
    \item \textbf{Privacy metrics}: Distance to Closest Record (DCR), membership inference attack success rate, and attribute inference risk, providing quantitative privacy assessments.
    \item \textbf{Diversity}: Coverage metrics measuring the fraction of real data modes represented in the synthetic data, and nearest-neighbor diversity ratios.
\end{itemize}

The evaluation module generates standardized reports enabling fair comparison across generative paradigms within a single experimental run.

%%=============================================================================
\section{Experiments}\label{sec:experiments}
%%=============================================================================

All experiments in this section are built on top of the proposed framework (Section~\ref{sec:framework}), using the GAN backend with the CTGAN architecture. The complete implementation, including data loading, preprocessing, model training, adversarial augmentation, and evaluation, is available at \url{https://github.com/Diyago/Tabular-data-generation}.

\subsection{Task Formulation}\label{sec:task}

Consider training and test sets $T_{\text{train}}$ and $T_{\text{test}}$ drawn from potentially different distributions. The goal is to improve predictive performance on $T_{\text{test}}$ by augmenting $T_{\text{train}}$ with GAN-generated synthetic data $T_{\text{synth}}$ that better approximates the test distribution, without using test labels.

\subsection{Experimental Design}\label{sec:exp_design}

\begin{figure*}[t]
\centering
\includegraphics[width=0.8\textwidth]{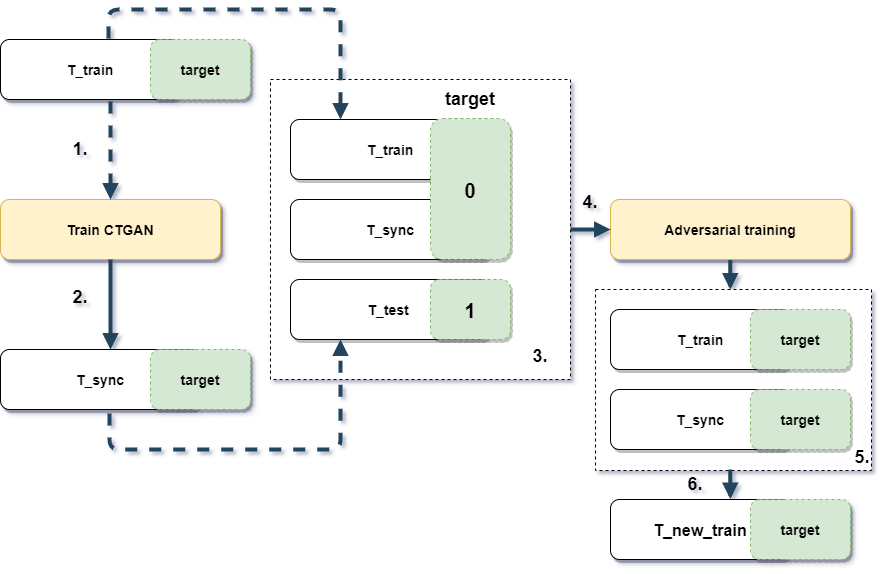}
\caption{Experimental pipeline implemented within the proposed framework. Steps 1--2 use the preprocessing and GAN modules; steps 3--5 use the training pipeline with adversarial augmentation; step 6 uses the evaluation module.}
\label{exp}
\end{figure*}

The experimental pipeline, illustrated in Figure~\ref{exp}, proceeds as follows:
\begin{enumerate}
    \item Train CTGAN on $T_{\text{train}}$ with ground-truth labels (\emph{framework preprocessing + GAN backend}).
    \item Generate synthetic data $T_{\text{synth}}$ (\emph{framework sampling}).
    \item Train a gradient boosting classifier in an adversarial manner on $T_{\text{train}} \cup T_{\text{synth}}$ (labeled 0) vs.\ $T_{\text{test}}$ (labeled 1), using only feature columns---no ground-truth labels from $T_{\text{test}}$ are used.
    \item Score all rows in $T_{\text{train}} \cup T_{\text{synth}}$ by predicted probability of belonging to $T_{\text{test}}$.
    \item Select top-scoring rows and train the final classifier on this filtered set.
    \item Evaluate on $T_{\text{test}}$ (\emph{framework evaluation module}).
\end{enumerate}

Three configurations are compared: (a)~\textbf{None}---training on unaugmented $T_{\text{train}}$; (b)~\textbf{GAN}---augmentation with CTGAN-generated data followed by adversarial filtering; (c)~\textbf{Sample Original}---adversarial filtering applied to $T_{\text{train}}$ without synthetic augmentation.

\subsection{Datasets}\label{sec:datasets}

Seven datasets from diverse domains are used, all targeting binary classification. Preprocessing removes time-based columns; remaining columns are either categorical or numerical. To study the effect of limited training data, subsets of varying sizes (5\%, 10\%, 25\%, 50\%, 75\%) are sampled from $T_{\text{train}}$. Dataset characteristics are summarized in Table~\ref{tbdatasets}.

\begin{table}[t]
\centering
\caption{Dataset characteristics.}
\label{tbdatasets}
\begin{tabular}{l|r|r|r|r|r}
\toprule
\textbf{Name} & \textbf{Total} & \textbf{Train} & \textbf{Test} & \textbf{Features} & \textbf{Cat.\ feat.} \\
\midrule
Telecom       & 7.0k   & 4.2k   & 2.8k   & 20  & 16 \\
Adult Income  & 48.8k  & 29.3k  & 19.5k  & 15  & 8  \\
Employee      & 32.7k  & 19.6k  & 13.1k  & 10  & 9  \\
Credit        & 307.5k & 184.5k & 123k   & 121 & 18 \\
Mortgages     & 45.6k  & 27.4k  & 18.2k  & 20  & 9  \\
Taxi          & 892.5k & 535.5k & 357k   & 8   & 5  \\
Poverty       & 37.6k  & 22.5k  & 15.0k  & 41  & 38 \\
\bottomrule
\end{tabular}
\end{table}

\subsection{Results}\label{sec:results}

Table~\ref{tbdatasetres} reports the best ROC~AUC score achieved by each augmentation strategy per dataset (scaled to percentage of maximum per-dataset AUC).

\begin{table}[!htbp]
\centering
\caption{Best ROC AUC by augmentation strategy per dataset (as fraction of per-dataset maximum). Bold indicates best per row.}
\label{tbdatasetres}
\resizebox{.95\columnwidth}{!}{
\begin{tabular}{l|c|c|c}
\toprule
\textbf{Dataset} & \textbf{None} & \textbf{GAN} & \textbf{Sample Original} \\
\midrule
Credit     & 0.997          & \textbf{0.998} & 0.997 \\
Employee   & \textbf{0.986} & 0.966          & 0.972 \\
Mortgages  & 0.984          & 0.964          & \textbf{0.988} \\
Poverty    & 0.937          & \textbf{0.950} & 0.933 \\
Taxi       & 0.966          & 0.938          & \textbf{0.987} \\
Adult      & 0.995          & 0.967          & \textbf{0.998} \\
Telecom    & \textbf{0.995} & 0.868          & 0.992 \\
\bottomrule
\end{tabular}
}
\end{table}

\begin{table}[!htbp]
\centering
\caption{Aggregated results across datasets. Higher mean AUC and lower std are preferred.}
\label{tbsamplers}
\begin{tabular}{l|c|c}
\toprule
\textbf{Strategy} & \textbf{Mean AUC} & \textbf{Std} \\
\midrule
None            & 0.980          & 0.036 \\
GAN             & 0.969          & 0.060 \\
Sample Original & \textbf{0.981} & \textbf{0.032} \\
\bottomrule
\end{tabular}
\end{table}

GAN-based augmentation achieves the best score on 2 of 7 datasets (Credit, Poverty), while Sample Original leads on 3 of 7 (Table~\ref{tbdatasetres}). In aggregate (Table~\ref{tbsamplers}), the three strategies perform comparably in mean AUC.

A more revealing analysis considers distribution shift. Let $\textit{same\_target}=1$ when the class ratio between train and test differs by no more than 5\%. As shown in Table~\ref{tbresults}, when distributions are similar ($\textit{same\_target}=1$), the baseline and Sample Original perform best. However, when distribution shift is present ($\textit{same\_target}=0$), GAN-based augmentation yields a higher AUC (0.966) than the unaugmented baseline (0.964), suggesting its particular utility under distributional mismatch.

\begin{table}[!htbp]
\centering
\caption{Performance stratified by distribution similarity between train and test sets.}
\label{tbresults}
\begin{tabular}{l|c|c}
\toprule
\textbf{Strategy} & \textbf{Same target} & \textbf{AUC} \\
\midrule
None            & 0 & 0.964 \\
None            & 1 & 0.985 \\
GAN             & 0 & 0.966 \\
GAN             & 1 & 0.945 \\
Sample Original & 0 & 0.973 \\
Sample Original & 1 & 0.984 \\
\bottomrule
\end{tabular}
\end{table}

\subsection{Reproducibility Statement}\label{sec:reproducibility}

All experiments are fully reproducible using the proposed framework's reference implementation at \url{https://github.com/Diyago/Tabular-data-generation}. The repository includes data loading scripts, preprocessing configurations, model training code, evaluation pipelines, and instructions for reproducing the reported results. Fixed random seeds and versioned dependencies ensure deterministic execution.

%%=============================================================================
\section{Discussion}\label{sec:discussion}
%%=============================================================================

\subsection{Data Quality Assessment}\label{sec:data_quality}

Evaluating synthetic tabular data quality requires multiple complementary perspectives. \emph{Statistical fidelity} measures how well the synthetic data matches the marginal and joint distributions of the original data. \emph{Machine learning utility}, typically assessed via the Train-on-Synthetic, Test-on-Real (TSTR) protocol, measures whether downstream models trained on synthetic data achieve comparable performance to those trained on real data. Our framework's evaluation module (Section~\ref{sec:evaluation}) implements both perspectives, enabling systematic quality assessment across generative paradigms.

Recent work has highlighted that aggregate metrics can mask column-level quality variations~\cite{borisov2022deep}. Our framework addresses this by providing per-column diagnostic reports alongside aggregate scores, allowing practitioners to identify specific failure modes (e.g., poor tail behavior in skewed numerical columns, or missing rare categories).

\subsection{Privacy Considerations}\label{sec:privacy}

Synthetic data is often motivated by privacy preservation, but generative models can memorize and reproduce training records~\cite{carlini2023extracting}. Key risks include:
\begin{itemize}
    \item \textbf{Membership inference}: Determining whether a specific record was in the training data.
    \item \textbf{Attribute inference}: Predicting sensitive attributes of a known individual from partially known features.
    \item \textbf{Data extraction}: Directly recovering training records from model outputs.
\end{itemize}

Diffusion models and GANs can both be augmented with differential privacy (DP) guarantees~\cite{abadi2016deep}, though this typically degrades sample quality. LLM-based generators face additional risks from pre-training data leakage. Our framework's evaluation module includes privacy metrics (DCR, membership inference success rate) to quantify these risks, and the pipeline supports DP-SGD training as a configurable option.

\subsection{Controllability}\label{sec:controllability}

Controllable generation---producing synthetic data satisfying user-specified constraints (e.g., class balance, feature ranges, conditional distributions)---is increasingly important for practical applications. GANs support limited controllability through conditional generation (as in CTGAN). Diffusion models enable controllability through classifier-free guidance~\cite{ho2022classifierfree}. LLMs offer the most natural controllability through prompt engineering and instruction following.

Our framework unifies these mechanisms through the \texttt{conditions} parameter in the \texttt{sample()} interface, abstracting paradigm-specific conditioning mechanisms behind a common API.

\subsection{Scalability}\label{sec:scalability}

Scalability concerns differ across paradigms:
\begin{itemize}
    \item \textbf{GANs} scale well to large datasets but may struggle with high-dimensional feature spaces due to mode collapse.
    \item \textbf{Diffusion models} incur quadratic memory cost in the number of features when using attention-based denoisers, and linear cost in the number of diffusion steps during inference.
    \item \textbf{LLMs} face token-length limitations that constrain the number of columns per row and require substantial GPU memory for fine-tuning.
\end{itemize}

The framework's modular design allows practitioners to select the appropriate paradigm based on dataset characteristics: GANs for large, low-dimensional datasets; diffusion models for complex distributions requiring high fidelity; and LLMs for semantically rich, few-shot scenarios.

%%=============================================================================
\section{Conclusion}\label{sec:conclusion}
%%=============================================================================

We have presented a comprehensive survey of tabular data generation spanning GANs, diffusion models, and large language models, and introduced a unified, modular framework that supports all three paradigms within a single experimental pipeline. The framework---comprising data preprocessing, model interface, training/inference pipeline, and evaluation modules---enables fair comparison, combination, and deployment of heterogeneous generative approaches.

Our experiments on seven benchmark datasets demonstrate that GAN-based augmentation, implemented within the framework, can improve downstream classification performance under distribution shift. The framework and its reference implementation (\url{https://github.com/Diyago/Tabular-data-generation}) are released as open-source software to support reproducibility and to serve as an extensible foundation for future research in tabular data generation.

Future directions include: (a)~integrating flow matching and energy-based generative models into the framework, (b)~extending support to multi-table relational data generation, (c)~incorporating differential privacy mechanisms as first-class components, and (d)~conducting large-scale benchmarks comparing all supported paradigms across diverse domains.

%%=============================================================================
\section*{Acknowledgments}
%%=============================================================================

The author would like to thank the Open Data Science community~\cite{ods} for valuable discussions and educational support.

%
% ---- Bibliography ----
%
\bibliographystyle{splncs04}
\bibliography{salt}
\end{document}